\documentclass[10pt, a4paper]{article}

\usepackage{lrec-coling2024} % this is the new style

\usepackage{microtype}
\usepackage{multirow}
\usepackage{subcaption}
\usepackage{amssymb}
\usepackage{amsmath}
\usepackage{enumitem}
\usepackage{array}
\usepackage{longtable}
\usepackage{makecell}
\usepackage{tabularx}
\usepackage{booktabs}

\usepackage{inconsolata}
\usepackage{pifont}

\definecolor{amber}{RGB}{245,188,20}
\definecolor{quantity}{RGB}{154,1,254}
\definecolor{gold}{RGB}{245,188,20}
\definecolor{scale}{RGB}{187,239,149}

\newcommand{\figref}[1]{Figure~\ref{fig:#1}}
\newcommand{\tabref}[1]{Table~\ref{tab:#1}}

\newcommand{\tatdqa}{\text{TAT-DQA}}
\newcommand{\model}{\text{Doc2SoarGraph}}

\newcommand{\llm}{LayoutLMv2$_\text{LARGE}$}
\newcommand\fone{F\textsubscript{1}}

\title{Doc2SoarGraph: Discrete Reasoning over Visually-Rich Table-Text Documents via Semantic-Oriented Hierarchical Graphs}

\name{Fengbin Zhu\textsuperscript{1},~Chao 
 Wang\textsuperscript{2},~Fuli Feng\textsuperscript{3*}$\thanks{*Corresponding author}$,~Zifeng Ren\textsuperscript{1}, Moxin Li\textsuperscript{1},~\textbf{Tat-Seng Chua}\textsuperscript{1}
\\
}
\address{\textsuperscript{1}National University of Singapore,~\textsuperscript{2}6Estates Pte Ltd,~\textsuperscript{3}University of Science and Technology of China\\
\tt{\{zhfengbin, fulifeng93\}@gmail.com}, \tt{wangchao@6estates.com}}

\abstract{
Table-text document understanding (e.g., financial reports) has attracted increasing attention in recent two years.
TAT-DQA~\citeplanguageresource{zhu2022towards} is a realistic setting for the understanding of visually-rich table-text documents, which involves answering associated questions requiring discrete reasoning. 
Most existing work relies on token-level semantics, falling short in the reasoning across document elements such as quantities and dates. 
To address this limitation, we propose a novel \textbf{\model}~model that exploits element-level semantics and employs {\bf S}emantic-{\bf o}riented hier{\bf ar}chical {\bf Graph} structures to capture the differences and correlations among different elements within the given document and question.
Extensive experiments on the TAT-DQA dataset reveal that our model surpasses the state-of-the-art conventional method (i.e., MHST) and large language model (i.e., ChatGPT) by $17.73$ and $6.49$ points respectively in terms of Exact Match (EM) metric, demonstrating exceptional effectiveness.
The source code is publicly available at  ~\url{https://github.com/fengbinzhu/Doc2SoarGraph/}.
\\ \newline  }

\begin{document}

\maketitleabstract

\section{Introduction}
Table-text documents containing a hybrid of tabular and textual data are pervasive in the real world, e.g. SEC filings, academic papers and medical reports.
Recently, there has been a surge of work attempting to intelligently understand table-text documents through answering associated questions~\citeplanguageresource{chen2020hybridqa,zhu2021tat,chen2021finqa,zhao2022multihiertt}.
However, these works focus on the well-annotated structured tables and manually selected paragraphs from the original documents, which is not in line with reality.

\begin{figure}[!t]
    % \toprule
    \begin{center}
    \includegraphics[scale=0.95
    ]{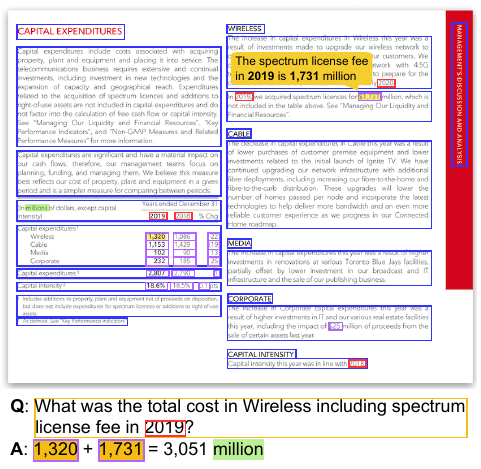}
    \end{center}
    \vspace{-0.7em}
    \caption{\label{fig:sample} 
    An example from \tatdqa. 
    We leverage four types of semantic elements from the question and document to facilitate discrete reasoning, i.e., \emph{Date}, \emph{Quantity}, \emph{Question} and \emph{Block}, marked  in \textcolor{red}{red}, \textcolor{quantity}{purple}, \textcolor{gold}{yellow} and \textcolor{blue}{blue} rectangle, respectively.
    The quantities with \colorbox{gold}{yellow} background are supporting evidence to the question.
    The ``million'' with \colorbox{scale}{green} background is the scale of the answer.
    }
    % \vspace{-2em}
\end{figure}

Research on the intelligent understanding of real-world table-text documents has been activated with the release of \tatdqa~\citeplanguageresource{zhu2022towards}, a Document Visual Question Answering (DocVQA) challenge over financial documents.
In \tatdqa, each document contains extensive numerical data in both tabular and textual formats, where discrete reasoning  capabilities (\textit{e.g., arithmetic calculation, comparison, counting and sorting}) are demanded to answer the questions.
One example is shown in ~\figref{sample}.
To address this challenge, MHST~\citep{zhu2022towards} applies sequence tagging on each token to select relevant tokens from the document, followed by answer inference over the selected tokens. 
Though effective, the performance of MHST is still not optimal.
One reason is that the tokens only carry part of the semantics of the original data.
%, resulting in sub-optimal performance.
For example, as shown in \figref{sample}, the spectrum license fee in~$2019$ is $1,731$ million, while the quantity $1,731$ corresponds to four tokens, i.e., ``$\underline{1}$'', ``$\underline{,}$'', ``$\underline{73}$'', and ``$\underline{\textrm{\#\#}1}$'' after tokenization. 
The model can hardly infer the meaning of the original number from every single token unless they are all combined.

Instead of only relying on token-level semantics, we propose to also exploit the element-level semantics to facilitate discrete reasoning. 
In particular, we consider four types of elements in the input, including \emph{Question}, \emph{Block}, \emph{Quantity} and \emph{Date}, as shown in \figref{sample}.
Specifically, \emph{Question} refers to the question; \emph{Block} refers to each document block generated by the OCR/PDF converter; \emph{Quantity}, \emph{Date} respectively refer to each quantity and each date in the question and the document block.
Each of these elements carries more complete semantics than single tokens that can be leveraged by the model.
The differences and correlations among them can provide rich and crucial clues for the model to conduct reasoning to derive the answer.
For example, though $2019$ and $1,731$ in \figref{sample} are both numerical values, actually the former refers to ``year 2019'' (date), while the latter is ``the spectrum license fee'' (quantity), so they cannot be compared. 
Obviously, it would be more appropriate to model the different types of elements separately. 
On the other hand, to understand the numerical value $1,731$ in the document, the model needs to consider the text information of the corresponding document block. 
Thus, the correlations of different elements should also be leveraged to facilitate the model's reasoning process.

In this work, we extend SoarGraph~\cite{Zhu2023SoarGraph} to tackle the challenge over visually-rich table-text documents in the form of \tatdqa~\cite{zhu2022towards}.
Specifically, we propose a \textbf{Doc2SoarGraph} model for question answering over visually-rich table-text \underline{\bf doc}uments with~\underline{\bf s}emantic-\underline{\bf o}riented hier\underline{\bf ar}chical \underline{\bf graph}s.
It models the differences and correlations of the elements (i.e., quantities, dates, question and document blocks) in the input data with hierarchy graph structures taking each element as one node, gaining more powerful discrete reasoning capabilities.
In TAT-DQA, about $20\%$ of the documents are multi-page. 
For a multi-page document, we first transform it to a single image of the model preferred dimension.
%with a simple yet effective method. 
Then, given a question and a document, we adopt LayoutLMv2~\cite{xu2021layoutlmv2} to take in the question, document text and the corresponding layout and document image, and initializes the representations of all semantic elements with the output.
After that, we construct a hierarchy of four graphs in two levels.
In the first level, we build three graphs: a Quantity Comparison (QC) graph to model the magnitude and comparison among all the \emph{Quantity} nodes; 
a Date Comparison (DC) graph to model the time sequence among all the \emph{Date} nodes;
a Text Relation (TR) graph with the \emph{Question} node and \emph{Block} nodes as these nodes usually contain rich text information.
In the second level, on top of these three graphs, a Semantic Dependency (SD) graph is built with all types of nodes to model the semantic relationships and dependencies among them.
Then, the model selects the most question-relevant nodes from the SD graph and applies different reasoning strategies over the selected nodes to derive the final answer based on the answer type.

Our main contributions are three-fold.
1) We propose to exploit element-level semantics to facilitate discrete reasoning over visually-rich table-text documents.
2) We develop a novel \model~model to model the differences and correlations among various elements with semantic-oriented hierarchical graph structures, which owns greatly enhanced evidence extraction and discrete reasoning capabilities.
3) We conduct extensive experiments on \tatdqa~dataset, and the results show that our \model~model outperforms both state-of-the-art conventional method (i.e., MHST) and large language model (LLMs) (i.e., ChatGPT) by $17.73$ and $6.46$ points respectively in Exact Match (EM), demonstrating remarkable effectiveness.

\section{\model~Model}
\begin{figure*}[t]
    \begin{center}
     \vspace{-1em}
    \includegraphics[scale=0.9]{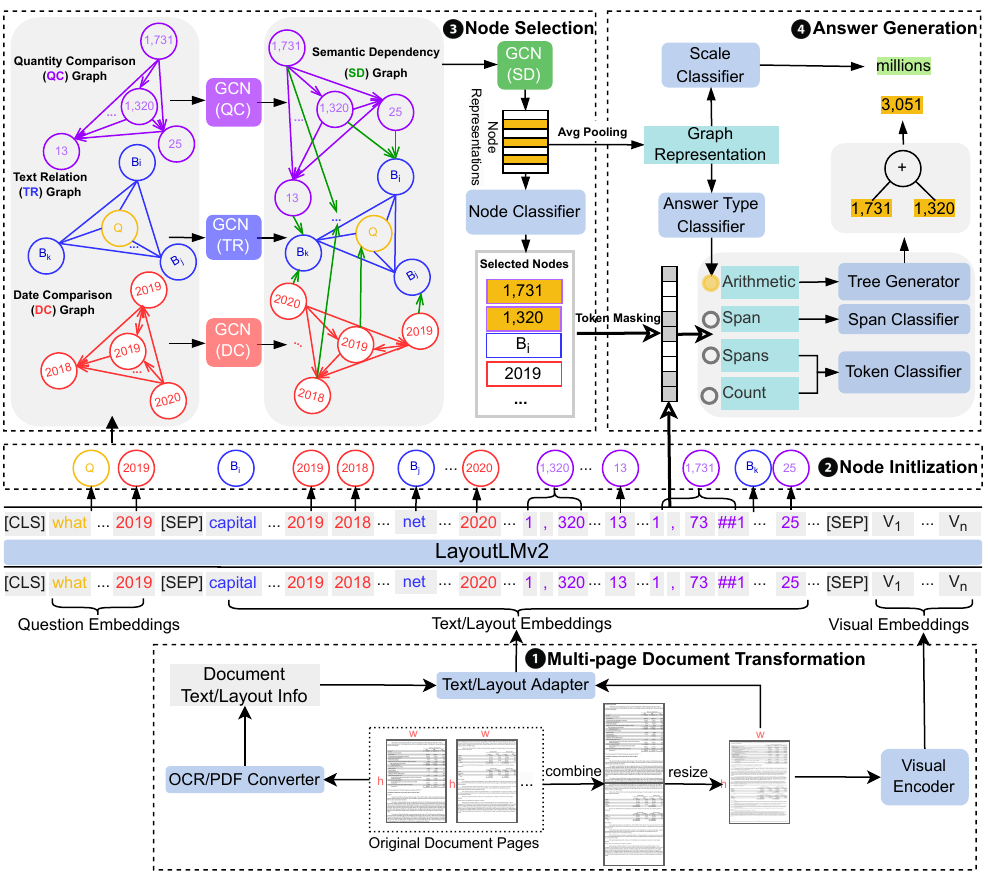}
    \end{center}
    % \vspace{-1em}
    \caption{\label{fig:model}
    An overview of proposed~\model~model. Take the sample in \figref{sample} as an example.}
     \vspace{-1em}
\end{figure*}

%We first formally define our problem. 
Consider a natural language question denoted as $Q$, and a visually-rich table-text document denoted as $D$ with several pages $P = (P_1, P_2, ..., P_{|P|}$),  where $|P|$ is the number of pages.
In the document $D$, the page $p$ has a list of blocks $B^{p} = (B^p_1, B^p_2, ..., B^p_{|B|})$ that are generated by an OCR/PDF converter, where $|B|$ is the number of blocks on the page $p$.
Our goal is to generate the answer to the question $Q$ that usually requires discrete reasoning based on the document $D$. 
To solve the problem, we develop a  \model~model.
An overall architecture is illustrated in \figref{model}.
%We elaborate the details of each component below.

\subsection{Document Transformation}
As pre-processing, we transform each multi-page document in TAT-DQA into a one-page document with a simple yet effective method. 
In particular, we first transform each page to a single image with the same dimension and then combine the corresponding multiple images of the pages vertically following the original page order.
Then, we resize the combined image to the dimension of a single-page document, which is preferred by the model.
Since the document text and layout information are available in \tatdqa, we further adjust the layout information according to the dimension of the final document image.
After that, all documents are considered as single-page documents and we obtain the initial visual embeddings of each document by applying the same CNN-based encoder as LayoutLMv2~\cite{xu2021layoutlmv2}.

\subsection{Node Initialization}
Rather than only relying on token-level semantics, our method also exploits element-level semantics to facilitate discrete reasoning with graph structures.
In particular, we harness four types of elements, namely, the question, each document block generated by the OCR/PDF converter, each quantity and each date in the question and the document block, which are named \emph{Question}, \emph{Block}, \emph{Quantity} and \emph{Date}, respectively.
We take each type of element as a kind of node, and get four types of nodes to build the graphs, i.e., \emph{Question} node, \emph{Block} node, \emph{Quantity} node and \emph{Date} node.
We then employ \llm~\cite{xu2021layoutlmv2} to take as input the question, the document text and layout information, and the final document image, and output the token-level hidden representations.
Then, we compute the mean of the corresponding tokens for each node as its initial representation.

\subsection{Node Selection}
\label{node_selection}
Based on the four types of nodes as explained above, we construct hierarchical graphs to model their relationships so as to select those most relevant nodes as the supporting evidence to the question and facilitate discrete reasoning of the model. 

\vspace{+0.6em}
\noindent $\bullet$ \textbf{Hierarchical Graphs Construction.}
\label{4graphs}
We construct four graphs, which form a two-level hierarchy, to model the element-level semantics.
Formally, a graph \textsc{G} is represented by an adjacency matrix $\textsc{A} \in R^{N \times N}$, where $N$ is the number of nodes.
If there is an edge connecting the $i^\textrm{th}$ and $j^\textrm{th}$ nodes, we assign value 1 to the corresponding position $(i, j)$ in the matrix $A$, and otherwise 0.  

\noindent \textbf{Quantity Comparison (QC) Graph} (denoted as $G_{QC}$): It is dedicated to retaining the numerical magnitude and comparison between every two quantities.
For two \emph{Quantity} nodes $q_i$, $q_j$, if $q_i$ $\geq$ $q_j$, a directed edge $e_{ij}$ = ($q_i$, $q_j$) pointing from $q_i$ to $q_j$ is added following NumNet~\cite{ran2019numnet}.

\noindent  \textbf{Date Comparison (DC) Graph} (denoted as $G_{DC}$): It is dedicated to retaining the time sequence and comparison between every two dates. 
For two \emph{Date} nodes $d_i$, $d_j$, a directed edge $e_{ij}$ = ($d_i$, $d_j$) pointing from $d_i$ to $d_j$ is added if $d_i$ $\geq$ $d_j$ ($d_i$ later than $d_j$).

\noindent  \textbf{Text Relation (TR) Graph} (denoted as $G_{TR}$):
It is dedicated to associating the informative descriptions among the question and the document blocks.
The \emph{Question} node and a \emph{Block} node or every two \emph{Block} nodes will have an undirected edge between them, forming a fully-connected graph. 

\noindent  \textbf{Semantic Dependency (SD) Graph} (denoted as $G_{SD}$):
It is built with all the four types of nodes to model the semantic dependencies of the \emph{Quantity} or \emph{Date} node upon the \emph{Question} or \emph{Block} node, besides attaining all the correlations in the above three graphs.
1) Edges for two \emph{Quantity} nodes, two \emph{Date} nodes, a \emph{Question} node and a block node, or two \emph{Block} nodes will be added in $G_{SD}$ following the construction rules for $G_{QC}$, $G_{DC}$, and $G_{TR}$, respectively. 2) Between one \emph{Quantity} node and one \emph{Question} or \emph{Block} node, a directed edge pointing from the \emph{Quantity} node to the \emph{Question} node or the \emph{Block} node will be added to the graph $G_{SD}$ if the quantity is part of the question or block; edges between one \emph{Date} node and one \emph{Question} or \emph{Block} node are added in the same way.

\vspace{+0.6em}
\noindent $\bullet$ \textbf{Node Classifier.}
\label{node_cls}
After constructing the hierarchical graphs, a dedicated graph convolution network (GCN)~\cite{kipf2017gcn} is applied for each graph to learn node representations respectively.
As illustrated in \figref{model}, the GCN (QC), GCN (DC) and GCN (TR) are applied respectively on the QC graph, DC graph and TR graph to learn corresponding node representations, which are then used to initialize the node representations of the SD graph. 
The GCN (SD) is applied on the SD graph to learn the final representation of each node $h_{node}$.
A binary node classifier is then applied on each node in the SD graph to predict whether the node is relevant to the question or not. 
The probability on node classification is computed as
\begin{align}
    \textrm{P}_{\text{node}} &= \textrm{softmax}(\textrm{FFN}(h_\textrm{node}))
\end{align}
where $\textrm{FFN}$ is a feed-forward network with two layers.
All the nodes that are classified as relevant to the question are collected.
The representation of the SD graph $h_{SD}$ is obtained by computing the mean of all the node representations in SD graph.

\subsection{Answer Generation}
We generate the final answer with the selected nodes, as follows. 
% \vspace{+0.6em}

\noindent $\bullet$ \textbf{Token Masking.}
\label{token_masking}
Based on the selected nodes, we mask the tokens that are not included in the selected \emph{Block} nodes to reduce the search space for answer prediction and update the token representations with their corresponding block node representations.
Particularly, we obtain the token-level representations from the output of the LayoutLMv2 encoder first.
Then, we mask the tokens not covered by any selected block nodes.
For tokens that are included in the selected block nodes, we update the representation of each token by concatenating its token representation with the corresponding block representation, 
\begin{align}
    h_{token}^{'}  &= \textrm{concat}(h_{token}, h_{node}) 
\end{align}
where $h_{token}$ is the token representation output from the encoder; 
$h_{node}$ is the representation of the token's corresponding block node obtained from the SD graph;
$concat$  denotes concatenation;
$h_{token}^{'}$ is the updated token representation.
For tokens that are masked in the sequence, we pad their representations with zero.
Finally, we obtain a sequence of updated token representations $h^{'}_{\text{[}t_1, t_2,...,t_s\text{]}}$ and $s$ is the maximum sequence length.

\vspace{+0.6em}
\noindent $\bullet$ \textbf{Answer Type Classifier.}
TAT-DQA offers four different answer types, i.e., \textit{Span}, \textit{Spans}, \textit{Counting}, \textit{Arithmetic}.
We adopt an Answer Type Classifier to predict the answer type of a question, which is essentially a multi-class classifier taking the SD graph representation $h_{SD}$ as input.
The probability of each answer type is computed as
\vspace{-0.3em}
\begin{align}
    \textrm{P}_{\text{type}} &= \textrm{softmax}(\textrm{FFN}(h_\textrm{SD})).
\end{align}
$\textrm{FFN}$ is a feed-forward network with two layers.

\vspace{+0.6em}
\noindent $\bullet$ \textbf{Span Classifier.}
For the \textit{Span} question, the answer is a sub-sequence of the input sequence, which is achieved by the Span Classifier. 
It takes the token representations obtained in Section \ref{token_masking} as the input and predicts the start and end indices of the sub-sequence. 
Formally, the probability distribution of the start position over the sequence is obtained by 
\begin{align}
 \textrm{P}_{\text{start}} &= \textrm{softmax}(\textrm{FFN}(h^{'}_{\text{[}t_1, t_2,...,t_s\text{]}})) 
\end{align}
where $\textrm{FFN}$ is a feed-forward network with two layers. 
Similarly, we can obtain the probability of the end position $\textrm{P}_{\text{end}}$.

\vspace{+0.6em}
\noindent $\bullet$ \textbf{Token Classifier.}
For the \textit{Spans} and \textit{Counting} questions, a Token Classifier is employed to infer the final answer.  
In particular, for each valid token obtained in Section~\ref{token_masking}, Token Classifier assigns a \texttt{B}, \texttt{I} or \texttt{O} label and takes those tagged with \texttt{B} and  \texttt{I} to generate the final answer. 
Formally, it takes in the updated representation $h_{token}^{'}$ of each valid token and computes the probability of the label as
\vspace{-0.3em}
\begin{align}
    \textrm{P}_\textrm{token} &= \textrm{softmax}(\textrm{FFN}({h_{token}^{'}}))
\end{align}
where $\textrm{FFN}$ is a feed-forward network with two layers.
After obtaining the tokens, the final answer for \emph{Spans} and \emph{Counting} questions is generated heuristically following MHST~\cite{zhu2022towards}.
\vspace{+0.6em}

\noindent $\bullet$ \textbf{Tree Generator.}
For the \textit{Arithmetic} question, a Tree Generator is adopted to generate an expression tree with the selected \emph{Quantity} and \emph{Date} nodes, which can be executed to infer the answer.
Following MHST~\cite{zhu2022towards}, the Tree Generator is implemented with GTS~\cite{xie2019goal}, which generates expression trees in a goal-driven manner.
To adapt GTS in our model, we make two major modifications.  
First, instead of feeding all the numbers and dates in the input into GTS, we only feed the selected most relevant \emph{Quantity} and \emph{Date} nodes, which significantly reduces the number of candidates for GTS to predict each leaf node and alleviates the difficulties. 
Second, when GTS predicts each node in the expression tree, we revise the generation of the context vector by attending to all the nodes in the SD graph instead of the tokens in the sequence, which can offer enhanced comprehensive semantic representations of the document.

The expression tree generated by the Tree Generator includes three kinds of nodes: the arithmetical operators $V_{op}$ (i.e., +,-,*,/), the constant numbers $V_{con}$ (i.e., $1$,$2$,$3$, .., $100$ ), and the quantity and date nodes $V_{node}$ that are selected in Section \ref{node_selection}.
The target vocabulary for tree generation is denoted as $V = V_{op} \cup V_{con} \cup V_{node}$ and its length is denoted as $L$.
Following the typical construction method of GTS~\cite{xie2019goal}, the expression tree is constructed starting from producing
the topmost operator and then the left and right child nodes.
\vspace{+0.6em}

\begin{table}[t]
    \centering
    \scriptsize
    \begin{center}
\begin{tabular}{llcc}
\toprule
\bf Type & \bf Model & \bf EM & \bf \fone \\
\midrule
\multicolumn{2}{l}{\bf Human Expert Performance} &  84.10 &	90.80 \\
% \midrule
\midrule
\multirow{4}{*}{\bf Fine-tuned } & 
NumNet+ V2 &  30.60 & 40.10\\
& TagOp & 33.70 & 42.50\\
& MHST & 41.50 & 50.70 \\
% \midrule
% \multirow{2}{*}{ \bf Ours} 
% & \multirow{2}{*}{\model} & (\textcolor{red}{+17.73})  & (\textcolor{red}{+16.91}) \\
% &  & \bf 59.20 & \bf 67.60 \\
% \midrule
\midrule
\multirow{4}{*}{\bf LLMs } 
& MAmmoTH (70B) & 35.42 &	42.82 \\
& WizardMath (70B)  & 36.44 &	41.55 \\
& LLaMA 2-Chat (70B) & 41.91 & 49.74 \\	
& ChatGPT & 52.74 & 61.40 \\
% & GPT-4 & \underline{64.46} & \underline{72.20} \\
\midrule
\multirow{2}{*}{ \bf Ours} 
& \multirow{2}{*}{\model} & (\textcolor{red}{+6.49})  & (\textcolor{red}{+6.21}) \\
&  & \bf 59.23 & \bf 67.61 \\
\bottomrule
\end{tabular}
\caption{Performance of our model and baseline models on the test set of TAT-DQA.}
    \label{tab:main}
    \vspace{-2.5em}
 \end{center}
\end{table}

\noindent $\bullet$ \textbf{Scale Classifier.}
Scale is vital for a numerical answer in TAT-DQA, including five possible values: \textit{Thousand}, \textit{Million}, \textit{Billion}, \textit{Percent} and \textit{None}.
A Scale Classifier is developed to predict the scale of the final answer.
In particular, it takes as input the SD graph representation $h_{SD}$ and computes the probability of each scale as
\vspace{-0.7em}
\begin{align}
    \textrm{P}_\textrm{scale} &= \textrm{softmax}(\textrm{FFN}(h_\textrm{SD}))
\end{align}
where $\textrm{FFN}$ is a feed-forward network with two layers.
After obtaining the scale, we generate the final answer by multiplying or concatenating the answer value with the scale following the practice in  MHST~\cite{zhu2022towards}.

\subsection{Training}
To optimize the proposed model, the objective is to minimize the sum of the losses of all classification tasks. 
Formally, the overall loss for each sample can be computed as
\begin{align}
\begin{split}
\mathcal{L} =& \mathcal{L}_{node}\textrm{+}\mathcal{L}_{tree}\textrm{+} \mathcal{L}_{start} \textrm{+}\mathcal{L}_{end} \textrm{+}  \mathcal{L}_{type}  \textrm{+} \mathcal{L}_{token} \textrm{+}  \\& \mathcal{L}_{scale} \\
\textrm{$\mathcal{L}_{node}$} &=~ \frac{1}{\mathopen|\mathrm{N}\mathclose|} \sum_{\mathrm{n} \in \mathrm{N}} \textrm{CE}(\textrm{P}_\textrm{node}^{\textrm{n}}, 
    \textrm{g}_\textrm{node}^{\textrm{n}})~ \\
\textrm{$\mathcal{L}_{tree}$} &=~ \frac{1}{\mathopen|\mathrm{S}\mathclose|} \sum_{\mathrm{s} \in \mathrm{S}} \textrm{CE}(\textrm{P}_(v^{s}|v^{1},...,v^{s-1},Q, G),  \textrm{g}^{s}_{v})~\\
    \textrm{$\mathcal{L}_{start}$} &=~ \textrm{CE}(\textrm{P}_{start}, \textrm{g}_{start})\\
    \textrm{$\mathcal{L}_{end}$} &=~ \textrm{CE}(\textrm{P}_{end}, \textrm{g}_{end})\\
    \textrm{$\mathcal{L}_{type}$} &=~ \textrm{CE}(\textrm{P}_{type}, \textrm{g}_{type})\\
   \textrm{$\mathcal{L}_{token}$} &=~ \frac{1}{\mathopen|\mathrm{T}\mathclose|} \sum_{\mathrm{t} \in \mathrm{T}} \textrm{CE}(\textrm{P}_\textrm{token}^{\textrm{t}}, 
    \textrm{g}_\textrm{token}^{\textrm{t}})~\\
    \textrm{$\mathcal{L}_{scale}$} &=~ \textrm{CE}(\textrm{P}_{scale}, \textrm{g}_{scale}).\\
\end{split}
\end{align}
Here $N$ is a set of nodes; $g_{node}^{n}$ is the ground-truth label if the node $n$ is selected; $\mathrm{CE}(\cdot)$ refers to the cross-entropy loss; $S$ is the number of decoding steps during the expression tree generation; $g_{v}^{s}$ is the ground-truth node in the step $s$; $g_{start}$ and $g_{end}$ are the ground-truth starting and ending positions of the span answer; $g_{type}$ is the ground-truth answer type; $T$ refers to all the valid tokens after applying the token masking in Section \ref{token_masking}; $g_{token}^{t}$ is the ground-truth label of the token $t$;
$g_{scale}$ is the ground-truth scale value.
When the nodes in the ground-truths in Node Selection are not selected, we will add them manually in order to better train the tree-based decoder when training.

\section{Experiments}
We conduct extensive experiments to validate the effectiveness of our proposed model and present comprehensive analyses.

% \subsection{Dataset, Baselines and Evaluation Metrics}
\subsection{Experiment Settings}

\noindent $\bullet$ \textbf{Dataset.}
We conduct all experiments on \tatdqa~\citeplanguageresource{zhu2022towards} built with visually-rich table-text documents in finance.
It contains $16,558$ QA pairs on $2,758$ documents where each document contains at least one table.
These documents are split into training, development and test sets with a ratio of $8:1:1$, and all the questions of a specific document belong to only one of the splits.
Over $50\%$ of questions require discrete reasoning to generate answers while the answers to others can be extracted directly from the documents.

\vspace{+0.6em}

\noindent $\bullet$ \textbf{Baselines.}
We select two kinds of baselines: fine-tuned models and large language models (LLMs).
\textbf{Fine-tuned models} are trained over \tatdqa~dataset, including:
1) NumNet+ V2~\cite{ran2019numnet} is a text QA model with impressive capability of discrete reasoning over textual data.
It constructs a numerically-aware graph neural network, which takes all numbers in the given question and document as nodes and builds edges via numerical comparison, and then performs discrete reasoning over the graph.
2) TagOp~\cite{zhu2021tat} is a table-text QA model which first applies sequence tagging on each token to extract question-relevant ones and then applies a set of pre-defined aggregation operators (e.g. addition, counting) over extracted tokens.
3) MHST~\cite{zhu2022towards} is a multi-modal QA model which employs LayoutLMv2~\cite{xu2021layoutlmv2} as the encoder to take the question and document as input, extracts supporting evidence using sequence tagging, and applies a tree-based decoder~\cite{xie2019goal} to generate an expression tree with the evidence.
\textbf{LLMs} are tested directly in a zero-shot manner, including two general LLMs, LaMA 2-Chat (70B) \cite{touvron2023llama2} and ChatGPT~\cite{openai2020gpt3}, and two LLMs specialized in math word problems (MWP), MAmmoTH (70B)~\cite{yue2023mammoth} and WizardMath (70B) 
\cite{luo2023wizardmath}.

\vspace{+0.6em}
\noindent $\bullet$ \textbf{Evaluation Metrics.}
Following~\cite{zhu2022towards}, Exact Match (EM) and numeracy-focused (macro-averaged) \fone{} are used to measure the performance of all models, taking into account the scale of the answer. Both metrics are in the range of [0\%, 100\%], where a higher value indicates better performance.

\vspace{+0.6em}

\noindent $\bullet$ \textbf{Implementation Details.}
We implement our model in PyTorch and train it on one NVIDIA DGX-1 with eight V100 GPUs.
We adopt LayoutLMv2$_{large}$ as the encoder.
We use Adam optimizer with learning rate $5e-4$ and warmup over the first $6\%$ steps to train. 
The maximum number of epochs is set to $50$ and the maximum sequence length $512$.
The batch size is set to $8$ and the number of gradient accumulation steps is $8$. 
The dropout probabilities for GCNs and GTS are $0.6$ and $0.5$ respectively while $0.1$ for others. 
We set $12$ as the maximum number of selected nodes in node selection. 
Beam search is applied during inference to select the best expression tree and the beam size is~$5$.

Given that the tabular and textual data in each document are typically lengthy, it is impractical to incorporate additional in-context examples owing to the input length constraints of the LLMs, thus we test all LLMs in a zero-shot setting. 
We utilize the latest ChatGPT\footnote{GPT3.5-Turbo (Sep 2023)}~\cite{openai2020gpt3} APIs\footnote{\url{https://platform.openai.com/}.}.
The parameters \texttt{temperature}, \texttt{top\_p} and \texttt{max\_tokens} are set with $0$, $1.0$ and $1,000$, and other parameters as default. 
We obtain the official trained checkpoints of LLaMA 2-Chat~\cite{touvron2023llama2}, MAmmoTH~\cite{yue2023mammoth} and WizardMath~\cite{luo2023wizardmath} from Hugginface\footnote{\url{https://huggingface.co/models}}.
The model inference is done on one NVIDIA DGX-A100 with eight A100 GPUs. 
The parameters \texttt{num\_beam} and \texttt{do\_sample} are 1 and false respectively.

\begin{table}[t]
    % \small
    \centering
\begin{tabular}{lrr|rr}
\toprule
\multirow{2}{*}{ Model } & \multicolumn{2}{c}{EM} &  \multicolumn{2}{c}{\fone{}} \\
 \cmidrule(lr){2-3}
 \cmidrule(lr){4-5}
& MHST & Ours & MHST & Ours \\
\midrule
Span & 41.10 & \bf 50.00 & 58.30 & \bf 62.88 \\
Spans & 25.70 & \bf  41.43 & 43.30 & \bf 71.19 \\
Counting & \bf 43.20 & 40.00 & \bf 43.20 & 40.00 \\
Arithmetic & 42.70 & \bf 73.96 & 42.70 & \bf 73.96 \\
\bottomrule
\end{tabular}
     \caption{Performance comparison of our model and MHST for different answer types on TAT-DQA test set. Best results are marked in bold.}
     \label{tab:answer-type}
       \vspace{-1.5em}
\end{table}

\subsection{Main Results}
We first compare our \model~model with all baseline models. 
The experimental results are shown in \tabref{main}.
We can observe that:
1) Our \model~model significantly outperforms all baseline models.
In particular, our model reaches $59.23\%$ and $67.61\%$ on the test set in terms of Exact Match and \fone{} metrics respectively, i.e., an increase of $17.73$ and $16.91$ points over MHST~\cite{zhu2022towards}, and $6.49$ and $6.21$ points over ChatGPT. 
These results well demonstrate the great effectiveness of our model.
2) The LLMs specialized in mathematical reasoning, i.e. WizardMath~\cite{luo2023wizardmath} and MAmmoTH~\cite{yue2023mammoth}, still largely underperform our \model~model, indicating that current numerically-enhanced LLMs still struggle in discrete reasoning over tabular and textual QA.
3) The best fine-tuned model MHST achieves comparable performance to the outstanding LLaMA-2 Chat with much smaller size, and \model~ largely outperforms the powerful ChatGPT. This shows that current general LLMs still struggle with table-text document QA, and fine-tuning on the dataset is still a promising approach.

\subsection{In-depth Analysis of Our Model}
\label{analysis}

\vspace{+0.6em}
\noindent $\bullet$ \textbf{Analysis on Evidence Extraction.}
\label{evidence-extraction}
Generally, for discrete reasoning over table-text documents, the model first extracts supporting evidence and then reasons over it.

\begin{figure}[t]
    \begin{center}
    \includegraphics[scale=0.54]{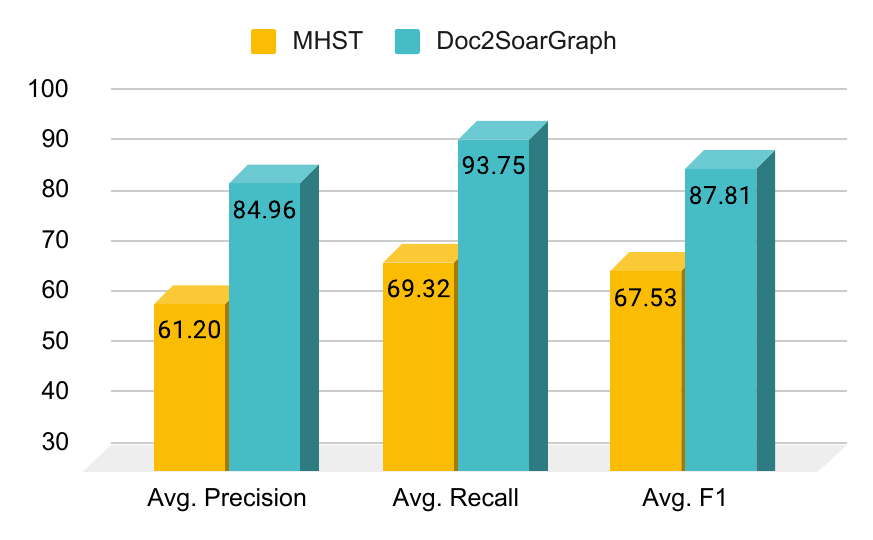}
    \caption{\label{fig:node_selection} 
    Comparison of evidence extraction power of our \model~and MHST on \textit{Arithmetic} questions on dev set.}
    \end{center}
    \vspace{-2em}
\end{figure}

Here we verify whether the evidence extraction power is indeed enhanced with our model.
We compute the average recall, precision and \fone{} score of the extracted evidence with our method and MHST on the dev set.
For fairness, we only use the \texttt{Arithmetic} questions that only depend on quantity and date nodes. 
Given one question, assume the number of quantities/dates its answer actually requires is $n$, the number of predicted quantities/dates by the model is $m$ and the number of correct quantities/dates in prediction is $c$.
The recall and precision are computed with $c/n$ and $c/m$ respectively.
Then we can compute the \fone{} with the precision and recall. 
After getting the metrics of each question, we further obtain the average recall, precision and \fone{}..
The results are shown in \figref{node_selection}.
We can see that our model demonstrates significant improvements over the MHST.
Specifically, our method has an increase of $23.76$ and $24.43$ points in average precision and average recall compared with MHST, significantly improving the evidence extraction.

\vspace{+0.6em}
\noindent $\bullet$ \textbf{Analysis on Answer Types.}
\tatdqa~provides four different answer types, and here we analyze the performance of our model on each answer type.
The results are summarized in \tabref{answer-type}.
Compared with MHST, our model gains the largest increase (i.e., $31.26$\% in EM) on \texttt{Arithmetic} questions, demonstrating impressive discrete reasoning capability.
This enhancement is possibly due to the effective modeling of the differences and correlations among the quantities, dates and blocks from the documents.
For \texttt{Spans} and \texttt{Counting} questions, they share almost all techniques in the proposed model.
Comparably, the model gains a $15.72$\% increase on \texttt{Spans} questions but has a $3.2$\% decrease on  \texttt{Counting} (i.e., failing one more case).
This is probably due to the data bias on \texttt{Counting} questions because the number of \texttt{Counting} questions (<$2.0$\%) is much less than \texttt{Spans} (>$12.0$\%) on test set.
The model obtains an increase of $8.9$\% in EM on \texttt{Span} questions, indicating our design also benefits answer extraction from the document.

\begin{table}[t]
    \scriptsize
    \centering
    \begin{tabular}{lrr}
   \toprule
      Model  & EM ($\uparrow$) &  \fone{} ($\uparrow$) \\
    \midrule
    
MHST &  41.50 (-) & 50.70 (-)   \\
\: + Node Initialization & 54.30 (12.80)  & 61.59 (10.89)  \\
\: + Doc Transformation  & 56.58 (2.28) & 64.06 (2.47) \\
\: + Hierarchical Graphs  &   58.80 (2.22) & 66.60 (2.54) \\
\: + Token Masking (Full)  &  \textbf{ 59.23} (0.43) &  \textbf{67.61} (1.06) \\
      \bottomrule
    \end{tabular}
    
     \caption{Analysis on effects of the components 
 in \model~on test set. Best results are marked in bold. }
    \label{tab:component}
    \vspace{-1em}
\end{table}

\begin{figure}[t]
    \begin{center}
    \includegraphics[scale=0.38]{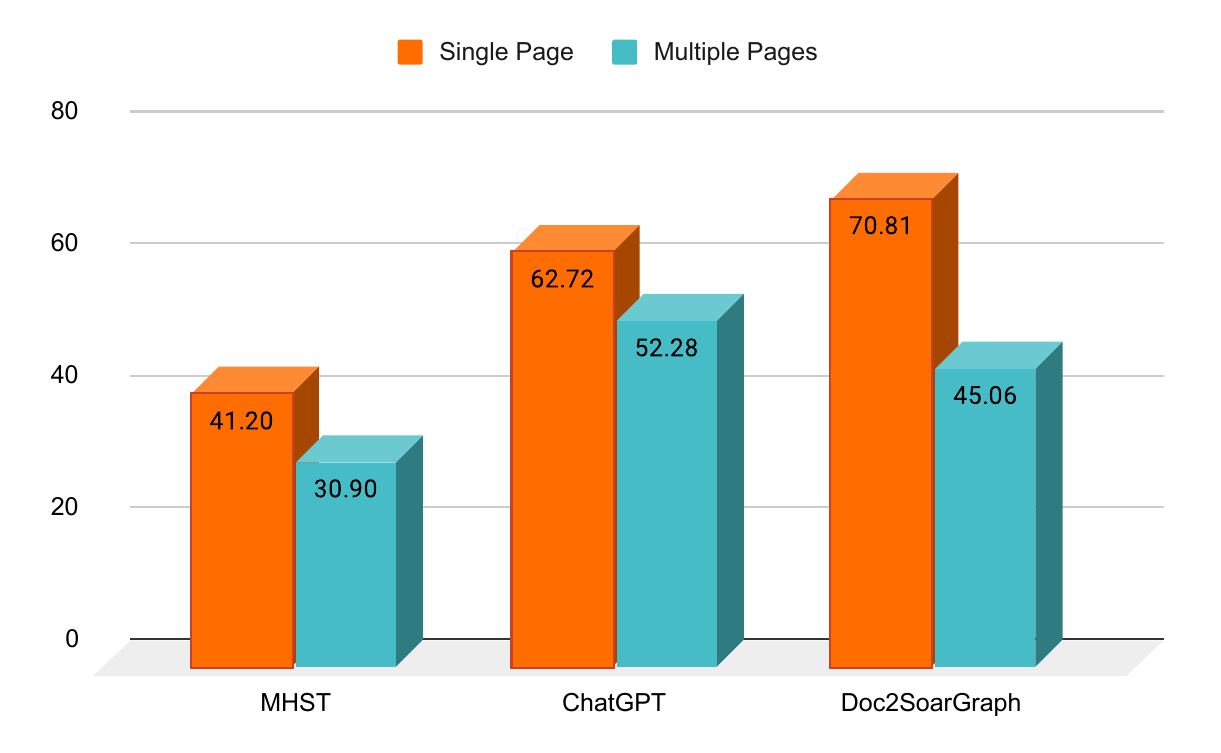}
     \caption{\label{fig:page-diff} 
    Performance comparison in \fone{} score on one- and multi-page documents on test set. 
    }
    \end{center}
    \vspace{-2em}
\end{figure}

\vspace{+0.6em}
\noindent $\bullet$ \textbf{Analysis on Single- and Multi-page Documents.}
\label{multi-page}
We analyze the performance differences of three models on single-page documents and multi-page documents, i.e. MHST, ChatGPT and our \model~model.
See \figref{page-diff} for the comparison results.  
We make following observations. 
1) All three models perform better on single-page documents than on multi-page ones, implying that it is more challenging to understand multi-page documents than single-page ones.
2) Our model outperforms MHST and ChatGPT with large margins for understanding single-page documents, further indicating the effectiveness of our model.
3) For multi-page document understanding, the performance of our model is better than MHST but worse than ChatGPT. 
The inferiority of our model to ChatGPT is mostly possibly due to the much shorter input allowed by our model.  

\begin{table}[t]
    \small
    \centering
    \begin{tabular}{lrrrr}
   \toprule
     \multirow{2}{*}{\bf Model}  & \multicolumn{2}{c}{\bf Dev} & \multicolumn{2}{c}{\bf Test} \\
     \cmidrule(lr){2-3}
     \cmidrule(lr){4-5}
        & EM & \fone{} & EM&  \fone{}\\
    \midrule
    Full Graphs & \bf 57.97 & \bf 65.38 & \bf 59.23 & \bf 67.61 \\
    \: - QC Graph & 56.69 & 65.18  & 57.73 & 66.89 \\
    \: - DC Graph & 56.14 & 64.83  & 57.73 & 66.82 \\
    \: - TR Graph & 55.23 & 63.57 & 55.86 & 65.09 \\
    \: - SD Graph & 56.27 & 64.77 & 56.95 & 66.54 \\
      \bottomrule
    \end{tabular}
     \caption{Ablation study of the hierarchical graphs in our model on test set.}
    \label{tab:ablation-graph}
     \vspace{-1.5em}
\end{table}

\vspace{+0.6em}
\noindent $\bullet$ \textbf{Analysis on Model Components.}
Our \model~contains four steps, i.e., Multi-page Document Transformation, Node Initialization according to various semantic elements, Node Selection via hierarchical graphs, and Answer Generation powered by token masking.
Here we investigate the contributions of each component to its final performance.
Compared to MHST, our~\model~ is equipped with the four components, and it is found that all added components can benefit model performance.
Furthermore, we find node initialization makes surprisingly greater contributions to the model, indicating the importance of modeling the differences and correlations among various elements in table-text documents.

Also, we develop our model with hierarchical graphs, i.e. the QC graph, DC graph, TR graph and SD graph.
To test the necessity of each graph, we remove each of them to see the performance changes. 
The results are summarized in \tabref{ablation-graph}.
Performance drop can be observed as we remove each graph, indicating that each graph contributes to the good performance of our model.

\vspace{+0.6em}
\noindent $\bullet$ \textbf{Error Analysis.}
\label{error-analysis}
We randomly sample $100$ error instances of our method on dev set and analyze the reasons.
We find the errors occur to all six modules (Col.~1 in \tabref{error-case}), i.e. Span Classifier (SPC), Token Classifier (TC), Node Classifier (NC), Tree Generator (TG), Scale Classifier (SC) and Answer Type Classifier (ATC), listed in a descending order of error percentage.
These errors are classified into nine categories (Col. 2 in \tabref{error-case}).
We can see, 
1) $32$\% errors are caused by SPC module predicting inaccurate predictions of starting and ending positions for \texttt{Span} questions, i.e., $21$\% predictions overlapping but not exactly matching ground truth, and $11$\% predictions having zero overlap with ground truth;
2) $24$\% errors are caused by NC module failing to select the relevant nodes;
3) $19$\% errors are due to TC module predicting less or more tokens than it needs to derive the answer; 
4) $16$\% errors are caused by TG module generating a wrong expression tree, among which $4$\% are wrong number signs (i.e., positive/negative) and $12$\% are other wrong expressions;
5) $6$\% and $3$\% errors are caused by SC module and ATC module predicting wrong scale and answer types.

\section{Related Work}
  
\vspace{+0.6em}
\subsection{Document VQA}
Document VQA aims to answer a question in natural language based on a visually-rich document~\cite{cui2021docai,Mathew2020DocVQA,VisualMRC2021,zhu2022towards}.
Compared to typical VQA, the documents in this challenges like DocVQA~\cite{Mathew2020DocVQA}, VisualMRC~\cite{VisualMRC2021} and TAT-DQA~\cite{zhu2022towards} usually contain rich textual information that plays a key role in addressing the challenge.
It is mostly tackled by pre-trained language models, e.g. LAMBERT~\cite{Garncarek2020LAMBERT}, StructuralLM~\cite{li2021structurallm} which exploit both textual and layout information of the documents.
 % multi-modal
Some works develop multi-modal language models that incorporate visual information into the model, e.g., LayoutLMv2~\cite{xu2021layoutlmv2} and DocFormer~\cite{Appalaraju2021DocFormer}.
Additionally, some DocVQA models are developed by fine-tuning pre-trained language models, e.g. TILT~\cite{powalski2021going} and MHST~\cite{zhu2022towards}.  
Recently, large-scale language models like ChatGPT~\cite{openai2020gpt3} have achieved impressive results across a range of natural language processing (NLP) tasks~\cite{zhao2023survey}.
In this work, we develop ~\model~ to comprehend visually-rich table-text documents, achieving comparable performance with the very large-scale language models like  ChatGPT~\cite{openai2020gpt3}.

% \vspace{+0.6em}

% \noindent $\bullet$ \textbf{Discrete Reasoning.}
\subsection{Discrete Reasoning}
Discrete reasoning has been explored in many NLP tasks since 1960s~\cite{feigenbaum1963computers,Dua2019DROP}.
Recent works focus on a hybrid of annotated (semi-)structured table and a list of associated paragraphs~\citeplanguageresource{chen2021finqa,  zhao2022multihiertt,zhu2021tat}, retrieving or extracting  evidences from given table and paragraphs and then reasoning over evidences to generate the answer \cite{lei2022answering,zhou2022unirpg,li2022finmath,nararatwong2022enhancing,Ramil2023num, zhu2021tat}.
Most recently, a document VQA dataset TAT-DQA~\cite{zhu2022towards} is released, which triggers the increasing interest in discrete reasoning over real-world complex documents with both tables and text.
To tackle this challenging task, \citet{zhu2022towards} proposed the MHST model, which extracts relevant tokens from the document using sequence tagging and applies heuristic or ``seq2tree'' method to generate the answer according to the answer type. 
We also address the TAT-DQA challenge but with a more powerful model.

\begin{table}[t]
\centering
\scriptsize
\begin{tabular} {l|l|r}
\toprule
Module & Error &  \% \\
\midrule
\multirow{2}{*}{Span Classifier (SPC) } & Offset Error & 21\% \\
\cmidrule{2-3}
& No Overlap & 11\% \\
\midrule
Node Classifier (NC) & Missing Nodes & 24\% \\
\midrule
\multirow{2}{*}{Token Classifier (TC) } & Missing Tokens & 11\% \\
\cmidrule{2-3}
& Redundant Tokens & 8\% \\
\midrule
\multirow{2}{*}{Tree Generator (TG) } & Wrong Expression & 12\% \\
\cmidrule{2-3}
& Wrong Sign & 4\% \\
\midrule
Scale Classifier (SC) & Wrong Scale & 6\% \\
\midrule
Answer Type Classifier (ATC) & Wrong Answer Type & 3\% \\
\bottomrule
\end{tabular} 

\caption{
 Statistics of errors in each module. }
\label{tab:error-case} 
% \vspace{-1.6em}
\end{table}

\subsection{Graph-based Document Representation}
% \noindent $\bullet$ \textbf{Graph-based Document Representation.}
Early works use grid-based methods to represent visually-rich documents, such as representing each document page as a grid of characters~\cite{katti2018chargrid} or a grid of contextualized word piece embedding vectors~\cite{denk2019bertgrid}.
% graph-based method 
Later, many works~\cite{Riba2019Table,hwang2021spatial,Wei2020Robust,Yu2020PICK,Cheng2020One} represent documents with more powerful graphs to facilitate information extraction from visually-rich documents.
For example,  \cite{Riba2019Table} adopts a GNN-based model to extract structured tables from invoices; 
% liu2019graph
\cite{hwang2021spatial} constructs a directed graph to model the spatial dependency among text tokens in the documents.
In this work, we represent the question and document by building hierarchical graphs with different semantic elements in the document (i.e., quantities, dates, and document blocks).
%to facilitate evidence extraction and discrete reasoning.

\section{Conclusion}
In this work, we propose a novel \model~model with strong discrete reasoning capabilities to tackle QA challenge over visually-rich table-text documents in the form of TAT-DQA, which models the differences and correlations of various elements (i.e., quantities, dates, question, and document blocks) in the input with hierarchical graphs.
We experimentally validate that our model can beat previous state-of-the-art by large margins.
In the future, we would like to explore more advanced methods to handle the challenging multi-page documents, even very long ones like financial statements with over 100 pages.

\section*{Limitations} 
Despite the impressive performance on TAT-DQA~\cite{zhu2022towards}, our model still has much room for future improvement, as shown in error analysis in Section \ref{error-analysis}. 
Also, the Document Transformation technique that we have developed for pre-processing multi-page documents is simple, and more effective methods are desired. 
For example, our model may not be applicable to the documents with a large number of pages (e.g., >50 pages).
In addition, our model is designed for the documents that contain different kinds of elements, such as numerical values and dates.
This means it may have limited advantages over those with unique elements like pure textual documents.

Furthermore,  our model has two major limitations on its discrete reasoning capabilities. 
First, the model is trained on TAT-DQA that is constructed with financial statements in the finance domain, which may result in limited applicability in other domains.
Second, the types of discrete reasoning supported are restricted to those in the benchmark, which currently includes operations such as addition, subtraction, multiplication, division, counting, comparison, sorting, and combinations.

\section*{Ethics Statement}
In this work, we present a new model to boost the  performance of discrete reasoning over visually-rich table-text documents.
Our model is developed on open-source tools and datasets to assist human-being in process .
Thus, we do not anticipate any negative social impacts.

\section*{Acknowledgements}
The authors gratefully thank all the anonymous reviewers for their positive feedback.
This research is supported by the NExT Research Center, Singapore.

\nocite{*}
\section{Bibliographical References}\label{sec:reference}

\bibliographystyle{lrec-coling2024-natbib}
\bibliography{lrec-coling2024-example}

\begin{thebibliography}{5}
\expandafter\ifx\csname natexlab\endcsname\relax\def\natexlab#1{#1}\fi

\bibitem[{Chen et~al.(2020)Chen, Zha, Chen, Xiong, Wang, and
  Wang}]{chen2020hybridqa}
Chen, Wenhu and Zha, Hanwen and Chen, Zhiyu and Xiong, Wenhan and Wang, Hong
  and Wang, William Yang. 2020.
\newblock \emph{{H}ybrid{QA}: A Dataset of Multi-Hop Question Answering over
  Tabular and Textual Data}.
\newblock Association for Computational Linguistics.

\bibitem[{Chen et~al.(2021)Chen, Chen, Smiley, Shah, Borova, Langdon, Moussa,
  Beane, Huang, Routledge, and Wang}]{chen2021finqa}
Chen, Zhiyu and Chen, Wenhu and Smiley, Charese and Shah, Sameena and Borova,
  Iana and Langdon, Dylan and Moussa, Reema and Beane, Matt and Huang, Ting-Hao
  and Routledge, Bryan and Wang, William Yang. 2021.
\newblock \emph{{F}in{QA}: A Dataset of Numerical Reasoning over Financial
  Data}.
\newblock Association for Computational Linguistics.

\bibitem[{Zhao et~al.(2022)Zhao, Li, Li, and Zhang}]{zhao2022multihiertt}
Zhao, Yilun and Li, Yunxiang and Li, Chenying and Zhang, Rui. 2022.
\newblock \emph{{M}ulti{H}iertt: Numerical Reasoning over Multi Hierarchical
  Tabular and Textual Data}.
\newblock Association for Computational Linguistics.

\bibitem[{Zhu et~al.(2022)Zhu, Lei, Feng, Wang, Zhang, and
  Chua}]{zhu2022towards}
Zhu, Fengbin and Lei, Wenqiang and Feng, Fuli and Wang, Chao and Zhang, Haozhou
  and Chua, Tat-Seng. 2022.
\newblock \emph{Towards complex document understanding by discrete reasoning}.

\bibitem[{Zhu et~al.(2021)Zhu, Lei, Huang, Wang, Zhang, Lv, Feng, and
  Chua}]{zhu2021tat}
Zhu, Fengbin and Lei, Wenqiang and Huang, Youcheng and Wang, Chao and Zhang,
  Shuo and Lv, Jiancheng and Feng, Fuli and Chua, Tat-Seng. 2021.
\newblock \emph{{TAT}-{QA}: A Question Answering Benchmark on a Hybrid of
  Tabular and Textual Content in Finance}.
\newblock Association for Computational Linguistics.

\end{thebibliography}


\begin{thebibliography}{66}
\expandafter\ifx\csname natexlab\endcsname\relax\def\natexlab#1{#1}\fi

\bibitem[{Andor et~al.(2019)Andor, He, Lee, and Pitler}]{andor2019giving}
Daniel Andor, Luheng He, Kenton Lee, and Emily Pitler. 2019.
\newblock Giving {BERT} a calculator: Finding operations and arguments with
  reading comprehension.
\newblock In \emph{EMNLP-IJCNLP}, pages 5947--5952. ACL.

\bibitem[{Appalaraju et~al.(2021)Appalaraju, Jasani, Kota, Xie, and
  Manmatha}]{Appalaraju2021DocFormer}
Srikar Appalaraju, Bhavan Jasani, Bhargava~Urala Kota, Yusheng Xie, and
  R.~Manmatha. 2021.
\newblock Docformer: End-to-end transformer for document understanding.
\newblock In \emph{Proceedings of the IEEE/CVF International Conference on
  Computer Vision (ICCV)}, pages 993--1003.

\bibitem[{Bobrow(1964)}]{bobrow1964natural}
Daniel~G Bobrow. 1964.
\newblock Natural language input for a computer problem solving system.

\bibitem[{Brown et~al.(2020)Brown, Mann, Ryder, Subbiah, Kaplan, Dhariwal,
  Neelakantan, Shyam, Sastry, Askell, Agarwal, Herbert-Voss, Krueger, Henighan,
  Child, Ramesh, Ziegler, Wu, Winter, Hesse, Chen, Sigler, Litwin, Gray, Chess,
  Clark, Berner, McCandlish, Radford, Sutskever, and Amodei}]{openai2020gpt3}
Tom Brown, Benjamin Mann, Nick Ryder, Melanie Subbiah, Jared~D Kaplan, Prafulla
  Dhariwal, Arvind Neelakantan, Pranav Shyam, Girish Sastry, Amanda Askell,
  Sandhini Agarwal, Ariel Herbert-Voss, Gretchen Krueger, Tom Henighan, Rewon
  Child, Aditya Ramesh, Daniel Ziegler, Jeffrey Wu, Clemens Winter, Chris
  Hesse, Mark Chen, Eric Sigler, Mateusz Litwin, Scott Gray, Benjamin Chess,
  Jack Clark, Christopher Berner, Sam McCandlish, Alec Radford, Ilya Sutskever,
  and Dario Amodei. 2020.
\newblock Language models are few-shot learners.
\newblock In \emph{Advances in Neural Information Processing Systems}, pages
  1877--1901.

\bibitem[{Chen et~al.(2020)Chen, Xu, Cheng, Xiaochuan, Zhang, Song, Wang, Qi,
  and Chu}]{chen2020question}
Kunlong Chen, Weidi Xu, Xingyi Cheng, Zou Xiaochuan, Yuyu Zhang, Le~Song,
  Taifeng Wang, Yuan Qi, and Wei Chu. 2020.
\newblock Question directed graph attention network for numerical reasoning
  over text.
\newblock In \emph{EMNLP-IJCNLP}, pages 6759--6768. ACL.

\bibitem[{Chen et~al.(2021)Chen, Zhao, Chen, Ji, Zhang, Luo, Xiong, and
  Yu}]{chen2021websrc}
Xingyu Chen, Zihan Zhao, Lu~Chen, JiaBao Ji, Danyang Zhang, Ao~Luo, Yuxuan
  Xiong, and Kai Yu. 2021.
\newblock {W}eb{SRC}: A dataset for web-based structural reading comprehension.
\newblock In \emph{Proceedings of the 2021 Conference on Empirical Methods in
  Natural Language Processing}, pages 4173--4185. Association for Computational
  Linguistics.

\bibitem[{Cheng et~al.(2020)Cheng, Qiu, Shi, Huang, and Lin}]{Cheng2020One}
Mengli Cheng, Minghui Qiu, Xing Shi, Jun Huang, and Wei Lin. 2020.
\newblock One-shot text field labeling using attention and belief propagation
  for structure information extraction.
\newblock In \emph{Proceedings of the 28th ACM International Conference on
  Multimedia}, MM '20, page 340–348. Association for Computing Machinery.

\bibitem[{Chiang and Chen(2019)}]{chen2019semantically}
Ting-Rui Chiang and Yun-Nung Chen. 2019.
\newblock Semantically-aligned equation generation for solving and reasoning
  math word problems.
\newblock In \emph{Proceedings of the 2019 Conference of the North {A}merican
  Chapter of the Association for Computational Linguistics: Human Language
  Technologies, Volume 1 (Long and Short Papers)}, pages 2656--2668.
  Association for Computational Linguistics.

\bibitem[{Cui et~al.(2021)Cui, Xu, Lv, and Wei}]{cui2021docai}
Lei Cui, Yiheng Xu, Tengchao Lv, and Furu Wei. 2021.
\newblock Document {AI:} benchmarks, models and applications.
\newblock \emph{CoRR}, abs/2111.08609.

\bibitem[{Denk and Reisswig(2019)}]{denk2019bertgrid}
Timo~I. Denk and Christian Reisswig. 2019.
\newblock {\{}BERT{\}}grid: Contextualized embedding for 2d document
  representation and understanding.
\newblock In \emph{Workshop on Document Intelligence at NeurIPS 2019}.

\bibitem[{Devlin et~al.(2019)Devlin, Chang, Lee, and
  Toutanova}]{Devlin2018Bert}
Jacob Devlin, Ming-Wei Chang, Kenton Lee, and Kristina Toutanova. 2019.
\newblock Bert: Pre-training of deep bidirectional transformers for language
  understanding.
\newblock In \emph{Proceedings of the 2019 Conference of the North American
  Chapter of the Association for Computational Linguistics: Human Language
  Technologies, Volume 1 (Long and Short Papers)}, pages 4171--4186.

\bibitem[{Dua et~al.(2019)Dua, Wang, Dasigi, Stanovsky, Singh, and
  Gardner}]{Dua2019DROP}
Dheeru Dua, Yizhong Wang, Pradeep Dasigi, Gabriel Stanovsky, Sameer Singh, and
  Matt Gardner. 2019.
\newblock {DROP}: A reading comprehension benchmark requiring discrete
  reasoning over paragraphs.
\newblock In \emph{Proc. of NAACL}.

\bibitem[{Feigenbaum et~al.(1963)Feigenbaum, Feldman
  et~al.}]{feigenbaum1963computers}
Edward~A Feigenbaum, Julian Feldman, et~al. 1963.
\newblock \emph{Computers and thought}.
\newblock New York McGraw-Hill.

\bibitem[{Garncarek et~al.(2020)Garncarek, Powalski, Stanislawek, Topolski,
  Halama, and Gralinski}]{Garncarek2020LAMBERT}
Lukasz Garncarek, Rafal Powalski, Tomasz Stanislawek, Bartosz Topolski, Piotr
  Halama, and Filip Gralinski. 2020.
\newblock \href {http://arxiv.org/abs/2002.08087} {{LAMBERT:} layout-aware
  language modeling using {BERT} for information extraction}.
\newblock \emph{CoRR}, abs/2002.08087.

\bibitem[{Herzig et~al.(2020)Herzig, Nowak, M{\"u}ller, Piccinno, and
  Eisenschlos}]{herzig2020tapas}
Jonathan Herzig, Pawel~Krzysztof Nowak, Thomas M{\"u}ller, Francesco Piccinno,
  and Julian Eisenschlos. 2020.
\newblock {T}a{P}as: Weakly supervised table parsing via pre-training.
\newblock In \emph{Proceedings of the 58th Annual Meeting of the Association
  for Computational Linguistics}, pages 4320--4333. ACL.

\bibitem[{Hong et~al.(2021)Hong, Kim, Ji, Hwang, Nam, and Park}]{hong2021bros}
Teakgyu Hong, DongHyun Kim, Mingi Ji, Wonseok Hwang, Daehyun Nam, and Sungrae
  Park. 2021.
\newblock \href {https://openreview.net/forum?id=punMXQEsPr0} {{\{}BROS{\}}: A
  pre-trained language model for understanding texts in document}.

\bibitem[{Hu et~al.(2019)Hu, Peng, Huang, and Li}]{hu2019multi}
Minghao Hu, Yuxing Peng, Zhen Huang, and Dongsheng Li. 2019.
\newblock A multi-type multi-span network for reading comprehension that
  requires discrete reasoning.
\newblock In \emph{Proceedings of the 2019 Conference on Empirical Methods in
  Natural Language Processing and the 9th International Joint Conference on
  Natural Language Processing (EMNLP-IJCNLP)}, pages 1596--1606. Association
  for Computational Linguistics.

\bibitem[{Huang et~al.(2017)Huang, Shi, Lin, and Yin}]{huang2017learning}
Danqing Huang, Shuming Shi, Chin-Yew Lin, and Jian Yin. 2017.
\newblock Learning fine-grained expressions to solve math word problems.
\newblock In \emph{Proceedings of the 2017 Conference on Empirical Methods in
  Natural Language Processing}, pages 805--814. ACL.

\bibitem[{Huang et~al.(2022)Huang, Lv, Cui, Lu, and Wei}]{huang2022layoutlmv3}
Yupan Huang, Tengchao Lv, Lei Cui, Yutong Lu, and Furu Wei. 2022.
\newblock \href {https://doi.org/10.1145/3503161.3548112} {Layoutlmv3:
  Pre-training for document ai with unified text and image masking}.
\newblock In \emph{Proceedings of the 30th ACM International Conference on
  Multimedia}, MM '22, page 4083–4091. Association for Computing Machinery.

\bibitem[{Hwang et~al.(2021)Hwang, Yim, Park, Yang, and Seo}]{hwang2021spatial}
Wonseok Hwang, Jinyeong Yim, Seunghyun Park, Sohee Yang, and Minjoon Seo. 2021.
\newblock Spatial dependency parsing for semi-structured document information
  extraction.
\newblock In \emph{Findings of the Association for Computational Linguistics:
  ACL-IJCNLP 2021}, pages 330--343. Association for Computational Linguistics.

\bibitem[{Jin et~al.(2022)Jin, Siebert, Li, and Chen}]{jin2022survey}
Nengzheng Jin, Joanna Siebert, Dongfang Li, and Qingcai Chen. 2022.
\newblock A survey on table question answering: Recent advances.
\newblock In \emph{China Conference on Knowledge Graph and Semantic Computing},
  pages 174--186. Springer.

\bibitem[{Katti et~al.(2018)Katti, Reisswig, Guder, Brarda, Bickel, H{\"o}hne,
  and Faddoul}]{katti2018chargrid}
Anoop~R Katti, Christian Reisswig, Cordula Guder, Sebastian Brarda, Steffen
  Bickel, Johannes H{\"o}hne, and Jean~Baptiste Faddoul. 2018.
\newblock {C}hargrid: Towards understanding 2{D} documents.
\newblock In \emph{Proceedings of the 2018 Conference on Empirical Methods in
  Natural Language Processing}, pages 4459--4469. Association for Computational
  Linguistics.

\bibitem[{Kipf and Welling(2017{\natexlab{a}})}]{kipf2017gcn}
Thomas~N. Kipf and Max Welling. 2017{\natexlab{a}}.
\newblock Semi-supervised classification with graph convolutional networks.
\newblock In \emph{International Conference on Learning Representations}.

\bibitem[{Kipf and Welling(2017{\natexlab{b}})}]{kipf2017semisupervised}
Thomas~N. Kipf and Max Welling. 2017{\natexlab{b}}.
\newblock Semi-supervised classification with graph convolutional networks.
\newblock In \emph{International Conference on Learning Representations}.

\bibitem[{Kushman et~al.(2014)Kushman, Artzi, Zettlemoyer, and
  Barzilay}]{kushman2014learning}
Nate Kushman, Yoav Artzi, Luke Zettlemoyer, and Regina Barzilay. 2014.
\newblock Learning to automatically solve algebra word problems.
\newblock In \emph{Proceedings of the 52nd Annual Meeting of the Association
  for Computational Linguistics}, pages 271--281. ACL.

\bibitem[{Lei et~al.(2022)Lei, He, Li, Zhao, and Liu}]{lei2022answering}
Fangyu Lei, Shizhu He, Xiang Li, Jun Zhao, and Kang Liu. 2022.
\newblock Answering numerical reasoning questions in table-text hybrid contents
  with graph-based encoder and tree-based decoder.
\newblock In \emph{Proceedings of the 29th International Conference on
  Computational Linguistics}, pages 1379--1390, Gyeongju, Republic of Korea.
  International Committee on Computational Linguistics.

\bibitem[{Li et~al.(2021{\natexlab{a}})Li, Bi, Yan, Wang, Huang, Huang, and
  Si}]{li2021structurallm}
Chenliang Li, Bin Bi, Ming Yan, Wei Wang, Songfang Huang, Fei Huang, and Luo
  Si. 2021{\natexlab{a}}.
\newblock {S}tructural{LM}: Structural pre-training for form understanding.
\newblock In \emph{Proceedings of the 59th Annual Meeting of the Association
  for Computational Linguistics and the 11th International Joint Conference on
  Natural Language Processing (Volume 1: Long Papers)}, pages 6309--6318.
  Association for Computational Linguistics.

\bibitem[{Li et~al.(2022{\natexlab{a}})Li, Ye, and Zhao}]{li2022finmath}
Chenying Li, Wenbo Ye, and Yilun Zhao. 2022{\natexlab{a}}.
\newblock Finmath: Injecting a tree-structured solver for question answering
  over financial reports.
\newblock In \emph{Proceedings of the Thirteenth Language Resources and
  Evaluation Conference}, pages 6147--6152.

\bibitem[{Li et~al.(2022{\natexlab{b}})Li, Feng, Zhang, He, Zhu, and
  Chua}]{li2022learning}
Moxin Li, Fuli Feng, Hanwang Zhang, Xiangnan He, Fengbin Zhu, and Tat-Seng
  Chua. 2022{\natexlab{b}}.
\newblock Learning to imagine: Integrating counterfactual thinking in neural
  discrete reasoning.
\newblock In \emph{Proceedings of the 60th Annual Meeting of the Association
  for Computational Linguistics (Volume 1: Long Papers)}, pages 57--69.
  Association for Computational Linguistics.

\bibitem[{Li et~al.(2021{\natexlab{b}})Li, Gu, Kuen, Morariu, Zhao, Jain,
  Manjunatha, and Liu}]{Li2021selfdoc}
Peizhao Li, Jiuxiang Gu, Jason Kuen, Vlad~I. Morariu, Handong Zhao, Rajiv Jain,
  Varun Manjunatha, and Hongfu Liu. 2021{\natexlab{b}}.
\newblock Selfdoc: Self-supervised document representation learning.
\newblock In \emph{Proceedings of the IEEE/CVF Conference on Computer Vision
  and Pattern Recognition (CVPR)}, pages 5652--5660.

\bibitem[{Lin et~al.(2021)Lin, Gao, Sun, Zhong, Hu, Ren, and
  Huo}]{Lin2021ViBertgrid}
Weihong Lin, Qifang Gao, Lei Sun, Zhuoyao Zhong, Kai Hu, Qin Ren, and Qiang
  Huo. 2021.
\newblock Vibertgrid: A jointly trained multi-modal 2d document representation
  for key information extraction from documents.
\newblock In \emph{Document Analysis and Recognition -- ICDAR 2021}, pages
  548--563. Springer International Publishing.

\bibitem[{Liu et~al.(2022)Liu, Piccinno, Krichene, Pang, Lee, Joshi, Altun,
  Collier, and Eisenschlos}]{liu2022matcha}
Fangyu Liu, Francesco Piccinno, Syrine Krichene, Chenxi Pang, Kenton Lee,
  Mandar Joshi, Yasemin Altun, Nigel Collier, and Julian~Martin Eisenschlos.
  2022.
\newblock Matcha: Enhancing visual language pretraining with math reasoning and
  chart derendering.
\newblock \emph{arXiv preprint arXiv:2212.09662}.

\bibitem[{Liu et~al.(2019{\natexlab{a}})Liu, Guan, Li, and
  Kawahara}]{liu2019tree}
Qianying Liu, Wenyv Guan, Sujian Li, and Daisuke Kawahara. 2019{\natexlab{a}}.
\newblock Tree-structured decoding for solving math word problems.
\newblock In \emph{Proceedings of the 2019 Conference on Empirical Methods in
  Natural Language Processing and the 9th International Joint Conference on
  Natural Language Processing (EMNLP-IJCNLP)}, pages 2370--2379. Association
  for Computational Linguistics.

\bibitem[{Liu et~al.(2019{\natexlab{b}})Liu, Gao, Zhang, and
  Zhao}]{liu2019graph}
Xiaojing Liu, Feiyu Gao, Qiong Zhang, and Huasha Zhao. 2019{\natexlab{b}}.
\newblock Graph convolution for multimodal information extraction from visually
  rich documents.
\newblock In \emph{Proceedings of the 2019 Conference of the North {A}merican
  Chapter of the Association for Computational Linguistics: Human Language
  Technologies, Volume 2 (Industry Papers)}, pages 32--39. Association for
  Computational Linguistics.

\bibitem[{Lu et~al.(2022)Lu, Qiu, Yu, Welleck, and Chang}]{lu2022survey}
Pan Lu, Liang Qiu, Wenhao Yu, Sean Welleck, and Kai-Wei Chang. 2022.
\newblock A survey of deep learning for mathematical reasoning.
\newblock \emph{arXiv preprint arXiv:2212.10535}.

\bibitem[{Luo et~al.(2023)Luo, Sun, Xu, Zhao, Lou, Tao, Geng, Lin, Chen, and
  Zhang}]{luo2023wizardmath}
Haipeng Luo, Qingfeng Sun, Can Xu, Pu~Zhao, Jianguang Lou, Chongyang Tao, Xiubo
  Geng, Qingwei Lin, Shifeng Chen, and Dongmei Zhang. 2023.
\newblock \href {http://arxiv.org/abs/2308.09583} {Wizardmath: Empowering
  mathematical reasoning for large language models via reinforced
  evol-instruct}.

\bibitem[{Mathew et~al.(2021)Mathew, Bagal, Tito, Karatzas, Valveny, and
  Jawahar}]{mathew2021infographicvqa}
Minesh Mathew, Viraj Bagal, Rubèn~Pérez Tito, Dimosthenis Karatzas, Ernest
  Valveny, and C.~V Jawahar. 2021.
\newblock \href {http://arxiv.org/abs/2104.12756} {Infographicvqa}.

\bibitem[{Mathew et~al.(2020)Mathew, Karatzas, Manmatha, and
  Jawahar}]{Mathew2020DocVQA}
Minesh Mathew, Dimosthenis Karatzas, R.~Manmatha, and C.~V. Jawahar. 2020.
\newblock \href {http://arxiv.org/abs/2007.00398} {Docvqa: {A} dataset for
  {VQA} on document images}.
\newblock \emph{CoRR}, abs/2007.00398.

\bibitem[{Mitra and Baral(2016)}]{mitra2016learning}
Arindam Mitra and Chitta Baral. 2016.
\newblock Learning to use formulas to solve simple arithmetic problems.
\newblock In \emph{Proceedings of the 54th Annual Meeting of the Association
  for Computational Linguistics}, pages 2144--2153. ACL.

\bibitem[{Nararatwong et~al.(2022)Nararatwong, Kertkeidkachorn, and
  Ichise}]{nararatwong2022enhancing}
Rungsiman Nararatwong, Natthawut Kertkeidkachorn, and Ryutaro Ichise. 2022.
\newblock Enhancing financial table and text question answering with tabular
  graph and numerical reasoning.
\newblock In \emph{Proceedings of the 2nd Conference of the Asia-Pacific
  Chapter of the Association for Computational Linguistics and the 12th
  International Joint Conference on Natural Language Processing (Volume 1: Long
  Papers)}, pages 991--1000. Association for Computational Linguistics.

\bibitem[{Pasupat and Liang(2015)}]{Pasupat2015Compositional}
Panupong Pasupat and Percy Liang. 2015.
\newblock Compositional semantic parsing on semi-structured tables.
\newblock In \emph{Proceedings of the 53rd Annual Meeting of the Association
  for Computational Linguistics and the 7th International Joint Conference on
  Natural Language Processing}, pages 1470--1480. ACL.

\bibitem[{Peng et~al.(2022)Peng, Pan, Wang, Luo, Zhang, Huang, Hu, Yin, Chen,
  Zhang et~al.}]{peng2022ernie}
Qiming Peng, Yinxu Pan, Wenjin Wang, Bin Luo, Zhenyu Zhang, Zhengjie Huang,
  Teng Hu, Weichong Yin, Yongfeng Chen, Yin Zhang, et~al. 2022.
\newblock Ernie-layout: Layout knowledge enhanced pre-training for
  visually-rich document understanding.
\newblock \emph{arXiv preprint arXiv:2210.06155}.

\bibitem[{Powalski et~al.(2021)Powalski, Borchmann, Jurkiewicz, Dwojak,
  Pietruszka, and Palka}]{powalski2021going}
Rafal Powalski, Lukasz Borchmann, Dawid Jurkiewicz, Tomasz Dwojak, Michal
  Pietruszka, and Gabriela Palka. 2021.
\newblock \href {http://arxiv.org/abs/2102.09550} {Going full-tilt boogie on
  document understanding with text-image-layout transformer}.
\newblock \emph{CoRR}, abs/2102.09550.

\bibitem[{Ran et~al.(2019)Ran, Lin, Li, Zhou, and Liu}]{ran2019numnet}
Qiu Ran, Yankai Lin, Peng Li, Jie Zhou, and Zhiyuan Liu. 2019.
\newblock {N}um{N}et: Machine reading comprehension with numerical reasoning.
\newblock In \emph{EMNLP-IJCNLP}, pages 2474--2484.

\bibitem[{Riba et~al.(2019)Riba, Dutta, Goldmann, Fornés, Ramos, and
  Lladós}]{Riba2019Table}
Pau Riba, Anjan Dutta, Lutz Goldmann, Alicia Fornés, Oriol Ramos, and Josep
  Lladós. 2019.
\newblock Table detection in invoice documents by graph neural networks.
\newblock In \emph{2019 International Conference on Document Analysis and
  Recognition (ICDAR)}, pages 122--127.

\bibitem[{Tanaka et~al.(2021)Tanaka, Nishida, and Yoshida}]{VisualMRC2021}
Ryota Tanaka, Kyosuke Nishida, and Sen Yoshida. 2021.
\newblock Visualmrc: Machine reading comprehension on document images.
\newblock In \emph{AAAI}.

\bibitem[{Tito et~al.(2022)Tito, Karatzas, and Valveny}]{Tito2022Hierarchical}
Rubèn Tito, Dimosthenis Karatzas, and Ernest Valveny. 2022.
\newblock Hierarchical multimodal transformers for multi-page docvqa.

\bibitem[{Touvron et~al.(2023)Touvron, Martin, Stone, Albert, Almahairi,
  Babaei, Bashlykov, Batra, Bhargava, Bhosale, Bikel, Blecher, Ferrer, Chen,
  Cucurull, Esiobu, Fernandes, Fu, Fu, Fuller, Gao, Goswami, Goyal, Hartshorn,
  Hosseini, Hou, Inan, Kardas, Kerkez, Khabsa, Kloumann, Korenev, Koura,
  Lachaux, Lavril, Lee, Liskovich, Lu, Mao, Martinet, Mihaylov, Mishra,
  Molybog, Nie, Poulton, Reizenstein, Rungta, Saladi, Schelten, Silva, Smith,
  Subramanian, Tan, Tang, Taylor, Williams, Kuan, Xu, Yan, Zarov, Zhang, Fan,
  Kambadur, Narang, Rodriguez, Stojnic, Edunov, and
  Scialom}]{touvron2023llama2}
Hugo Touvron, Louis Martin, Kevin Stone, Peter Albert, Amjad Almahairi, Yasmine
  Babaei, Nikolay Bashlykov, Soumya Batra, Prajjwal Bhargava, Shruti Bhosale,
  Dan Bikel, Lukas Blecher, Cristian~Canton Ferrer, Moya Chen, Guillem
  Cucurull, David Esiobu, Jude Fernandes, Jeremy Fu, Wenyin Fu, Brian Fuller,
  Cynthia Gao, Vedanuj Goswami, Naman Goyal, Anthony Hartshorn, Saghar
  Hosseini, Rui Hou, Hakan Inan, Marcin Kardas, Viktor Kerkez, Madian Khabsa,
  Isabel Kloumann, Artem Korenev, Punit~Singh Koura, Marie-Anne Lachaux,
  Thibaut Lavril, Jenya Lee, Diana Liskovich, Yinghai Lu, Yuning Mao, Xavier
  Martinet, Todor Mihaylov, Pushkar Mishra, Igor Molybog, Yixin Nie, Andrew
  Poulton, Jeremy Reizenstein, Rashi Rungta, Kalyan Saladi, Alan Schelten, Ruan
  Silva, Eric~Michael Smith, Ranjan Subramanian, Xiaoqing~Ellen Tan, Binh Tang,
  Ross Taylor, Adina Williams, Jian~Xiang Kuan, Puxin Xu, Zheng Yan, Iliyan
  Zarov, Yuchen Zhang, Angela Fan, Melanie Kambadur, Sharan Narang, Aurelien
  Rodriguez, Robert Stojnic, Sergey Edunov, and Thomas Scialom. 2023.
\newblock \href {http://arxiv.org/abs/2307.09288} {Llama 2: Open foundation and
  fine-tuned chat models}.

\bibitem[{Wang et~al.(2022{\natexlab{a}})Wang, Dou, and Che}]{wang2022survey}
Dingzirui Wang, Longxu Dou, and Wanxiang Che. 2022{\natexlab{a}}.
\newblock A survey on table-and-text hybridqa: Concepts, methods, challenges
  and future directions.
\newblock \emph{arXiv preprint arXiv:2212.13465}.

\bibitem[{Wang et~al.(2018)Wang, Wang, Cai, Zhang, and
  Liu}]{wang2018translating}
Lei Wang, Yan Wang, Deng Cai, Dongxiang Zhang, and Xiaojiang Liu. 2018.
\newblock Translating a math word problem to a expression tree.
\newblock In \emph{Proceedings of the 2018 Conference on Empirical Methods in
  Natural Language Processing}, pages 1064--1069. Association for Computational
  Linguistics.

\bibitem[{Wang et~al.(2022{\natexlab{b}})Wang, Huang, Luo, Chen, Peng, Pan,
  Yin, Feng, Sun, Yu, and Zhang}]{wang2022mmlayout}
Wenjin Wang, Zhengjie Huang, Bin Luo, Qianglong Chen, Qiming Peng, Yinxu Pan,
  Weichong Yin, Shikun Feng, Yu~Sun, Dianhai Yu, and Yin Zhang.
  2022{\natexlab{b}}.
\newblock \href {https://doi.org/10.1145/3503161.3548406} {Mmlayout:
  Multi-grained multimodal transformer for document understanding}.
\newblock MM '22, page 4877–4886. Association for Computing Machinery.

\bibitem[{Wang et~al.(2017)Wang, Liu, and Shi}]{wang2017deep}
Yan Wang, Xiaojiang Liu, and Shuming Shi. 2017.
\newblock Deep neural solver for math word problems.
\newblock In \emph{Proceedings of the 2017 Conference on Empirical Methods in
  Natural Language Processing}, pages 845--854. Association for Computational
  Linguistics.

\bibitem[{Wei et~al.(2020)Wei, He, and Zhang}]{Wei2020Robust}
Mengxi Wei, YIfan He, and Qiong Zhang. 2020.
\newblock Robust layout-aware ie for visually rich documents with pre-trained
  language models.
\newblock In \emph{Proceedings of the 43rd International ACM SIGIR Conference
  on Research and Development in Information Retrieval}, SIGIR '20, page
  2367–2376. Association for Computing Machinery.

\bibitem[{Xie and Sun(2019)}]{xie2019goal}
Zhipeng Xie and Shichao Sun. 2019.
\newblock A goal-driven tree-structured neural model for math word problems.
\newblock In \emph{IJCAI}, pages 5299--5305.

\bibitem[{Xu et~al.(2021)Xu, Xu, Lv, Cui, Wei, Wang, Lu, Florencio, Zhang, Che,
  Zhang, and Zhou}]{xu2021layoutlmv2}
Yang Xu, Yiheng Xu, Tengchao Lv, Lei Cui, Furu Wei, Guoxin Wang, Yijuan Lu,
  Dinei Florencio, Cha Zhang, Wanxiang Che, Min Zhang, and Lidong Zhou. 2021.
\newblock {L}ayout{LM}v2: Multi-modal pre-training for visually-rich document
  understanding.
\newblock In \emph{Proceedings of the 59th Annual Meeting of the Association
  for Computational Linguistics and the 11th International Joint Conference on
  Natural Language Processing (Volume 1: Long Papers)}, pages 2579--2591.
  Association for Computational Linguistics.

\bibitem[{Xu et~al.(2020)Xu, Li, Cui, Huang, Wei, and Zhou}]{xu2020layoutlm}
Yiheng Xu, Minghao Li, Lei Cui, Shaohan Huang, Furu Wei, and Ming Zhou. 2020.
\newblock Layoutlm: Pre-training of text and layout for document image
  understanding.
\newblock In \emph{Proceedings of the 26th ACM SIGKDD International Conference
  on Knowledge Discovery \& Data Mining}, page 1192–1200. Association for
  Computing Machinery.

\bibitem[{Yarullin and Isaev(2023)}]{Ramil2023num}
Ramil Yarullin and Sergei Isaev. 2023.
\newblock Numerical embeddings for reasoning over text and tables.

\bibitem[{Yin et~al.(2020)Yin, Neubig, Yih, and Riedel}]{yin2020tabert}
Pengcheng Yin, Graham Neubig, Wen-tau Yih, and Sebastian Riedel. 2020.
\newblock {T}a{BERT}: Pretraining for joint understanding of textual and
  tabular data.
\newblock In \emph{ACL}, pages 8413--8426. ACL.

\bibitem[{Yu et~al.(2020)Yu, Lu, Qi, Gong, and Xiao}]{Yu2020PICK}
Wenwen Yu, Ning Lu, Xianbiao Qi, Ping Gong, and Rong Xiao. 2020.
\newblock {PICK}: Processing key information extraction from documents using
  improved graph learning-convolutional networks.
\newblock In \emph{2020 25th International Conference on Pattern Recognition
  (ICPR)}.

\bibitem[{Yue et~al.(2023)Yue, Qu, Zhang, Fu, Huang, Sun, Su, and
  Chen}]{yue2023mammoth}
Xiang Yue, Xingwei Qu, Ge~Zhang, Yao Fu, Wenhao Huang, Huan Sun, Yu~Su, and
  Wenhu Chen. 2023.
\newblock \href {http://arxiv.org/abs/2309.05653} {Mammoth: Building math
  generalist models through hybrid instruction tuning}.

\bibitem[{Zhang et~al.(2020)Zhang, Wang, Lee, Bin, Wang, Shao, and
  Lim}]{zhang2020graphtree}
Jipeng Zhang, Lei Wang, Roy Ka-Wei Lee, Yi~Bin, Yan Wang, Jie Shao, and Ee-Peng
  Lim. 2020.
\newblock Graph-to-tree learning for solving math word problems.
\newblock In \emph{Proceedings of the 58th Annual Meeting of the Association
  for Computational Linguistics}, pages 3928--3937. Association for
  Computational Linguistics.

\bibitem[{Zhao et~al.(2023)Zhao, Zhou, Li, Tang, Wang, Hou, Min, Zhang, Zhang,
  Dong, Du, Yang, Chen, Chen, Jiang, Ren, Li, Tang, Liu, Liu, Nie, and
  Wen}]{zhao2023survey}
Wayne~Xin Zhao, Kun Zhou, Junyi Li, Tianyi Tang, Xiaolei Wang, Yupeng Hou,
  Yingqian Min, Beichen Zhang, Junjie Zhang, Zican Dong, Yifan Du, Chen Yang,
  Yushuo Chen, Zhipeng Chen, Jinhao Jiang, Ruiyang Ren, Yifan Li, Xinyu Tang,
  Zikang Liu, Peiyu Liu, Jian-Yun Nie, and Ji-Rong Wen. 2023.
\newblock \href {http://arxiv.org/abs/2303.18223} {A survey of large language
  models}.

\bibitem[{Zhou et~al.(2022)Zhou, Bao, Duan, Wu, He, and Zhao}]{zhou2022unirpg}
Yongwei Zhou, Junwei Bao, Chaoqun Duan, Youzheng Wu, Xiaodong He, and Tiejun
  Zhao. 2022.
\newblock Unirpg: Unified discrete reasoning over table and text as program
  generation.
\newblock \emph{arXiv preprint arXiv:2210.08249}.

\bibitem[{Zhu et~al.(2022)Zhu, Lei, Feng, Wang, Zhang, and
  Chua}]{zhu2022towards}
Fengbin Zhu, Wenqiang Lei, Fuli Feng, Chao Wang, Haozhou Zhang, and Tat-Seng
  Chua. 2022.
\newblock Towards complex document understanding by discrete reasoning.
\newblock In \emph{Proceedings of the 30th ACM International Conference on
  Multimedia}, pages 4857--4866.

\bibitem[{Zhu et~al.(2021)Zhu, Lei, Huang, Wang, Zhang, Lv, Feng, and
  Chua}]{zhu2021tat}
Fengbin Zhu, Wenqiang Lei, Youcheng Huang, Chao Wang, Shuo Zhang, Jiancheng Lv,
  Fuli Feng, and Tat-Seng Chua. 2021.
\newblock {TAT}-{QA}: A question answering benchmark on a hybrid of tabular and
  textual content in finance.
\newblock In \emph{Proceedings of the 59th Annual Meeting of the Association
  for Computational Linguistics and the 11th International Joint Conference on
  Natural Language Processing (Volume 1: Long Papers)}, pages 3277--3287.
  Association for Computational Linguistics.

\bibitem[{Zhu et~al.(2023)Zhu, Li, Xiao, Feng, Wang, and
  Chua}]{Zhu2023SoarGraph}
Fengbin Zhu, Moxin Li, Junbin Xiao, Fuli Feng, Chao Wang, and Tat~Seng Chua.
  2023.
\newblock Soargraph: Numerical reasoning over financial table-text data via
  semantic-oriented hierarchical graphs.
\newblock In \emph{Companion Proceedings of the ACM Web Conference 2023}, page
  1236–1244. Association for Computing Machinery.

\end{thebibliography}

\section{Language Resource References}
\label{lr:ref}
\bibliographystylelanguageresource{lrec-coling2024-natbib}
\bibliographylanguageresource{languageresource}

\end{document}